\def\year{2019}\relax
\documentclass[letterpaper]{article} %DO NOT CHANGE THIS
\usepackage{aaai19}  %Required
\usepackage{times}  %Required
\usepackage{helvet}  %Required
\usepackage{courier}  %Required
\usepackage{url}  %Required
\usepackage{graphicx}  %Required
 \usepackage{booktabs}
\usepackage[inline]{enumitem}
\usepackage[colorinlistoftodos]{todonotes}
\usepackage{subcaption}

\newcommand{\seeUrl}[1]{\footnote{\url{#1}}}

\newcommand*{\affaddr}[1]{#1} % No op here. Customize it for different styles.
\newcommand*{\affmark}[1][*]{\textsuperscript{#1}}
\newcommand*{\email}[1]{\texttt{#1}}

\begin{document}
% The file aaai.sty is the style file for AAAI Press 
% proceedings, working notes, and technical reports.
%
%\title{The Profiling Machine}
\title{The Profiling Machine: Active Generalization over Knowledge}

% \author{Filip Ilievski}\textsuperscript{1}
% \author{Eduard Hovy}\textsuperscript{2} 
% \author{Qizhe Xie}\textsuperscript{2}
% \author{Piek Vossen}\textsuperscript{1} \\
% \textsuperscript{1}{Computational Lexicology and Terminology Lab, Vrije Universiteit Amsterdam }\\
% \textsuperscript{2}{Language Technologies Institute, Carnegie Mellon University} \\
% \{f.ilievski, piek.vossen\}@vu.nl, 
% hovy@cmu.edu,
% qzxie@cs.cmu.edu

% \author{Filip Ilievski \\Computational Lexicology and Terminology Lab\\Vrije Universiteit Amsterdam \And Filip Ilievski \\Computational Lexicology and Terminology Lab\\Vrije Universiteit Amsterdam \And Filip Ilievski \\Computational Lexicology and Terminology Lab\\Vrije Universiteit Amsterdam \And Filip Ilievski \\Computational Lexicology and Terminology Lab\\Vrije Universiteit Amsterdam}

 \author{Filip Ilievski\affmark[1], Eduard Hovy\affmark[2], Qizhe Xie\affmark[2], Piek Vossen\affmark[1] \\
   \affaddr{\affmark[1]Computational Lexicology and Terminology Lab,
  Vrije Universiteit Amsterdam \\}
   \email{\{f.ilievski, piek.vossen\}@vu.nl}  \\
   \affaddr{\affmark[2]Language Technologies Institute,
   Carnegie Mellon University \\}
   \email{hovy@cmu.edu}, \email{qzxie@cs.cmu.edu}
%   \\https://www.overleaf.com/17643317cnyhxhjfhgmm
 }
%\author{...}
\maketitle

% \usepackage{graphicx}
% \DeclareGraphicsExtensions{.jpg,.mps,.pdf,.png}
% \usepackage{biblatex}
% \usepackage{listings}
% \usepackage[caption=false]{subfig}
% \usepackage{verbatim}
% \usepackage{xspace}
% %\usepackage{hyperref}

% \usepackage{placeins}
% \setlength{\floatsep}{4pt plus 0pt minus 2.0pt}
% \setlength{\textfloatsep}{4pt plus 0pt minus 2.0pt}

% \pagestyle{plain}

% \author{Filip Ilievski\affmark[1], Eduard Hovy\affmark[2], Qizhe Xie\affmark[2], Piek Vossen\affmark[1] \\
%   \affaddr{\affmark[1]Computational Lexicology and Terminology Lab,
%  Vrije Universiteit Amsterdam \\}
%   \email{\{f.ilievski, piek.vossen\}@vu.nl}  \\
%   \affaddr{\affmark[2]Language Technologies Institute,
%   Carnegie Mellon University \\}
%   \email{hovy@cmu.edu}, \email{qzxie@cs.cmu.edu}
%   }

\begin{abstract}
The human mind is a powerful multifunctional knowledge storage and management system that performs generalization, type inference, anomaly detection, stereotyping, and other tasks. A dynamic KR system that appropriately profiles over sparse inputs to provide complete expectations for unknown facets can help with all these tasks. 
%It is surprising that the quite natural challenge of profiling has not been addressed (widely) in the AI or NLP literature. 
%In this paper, we  which has not been addressed widely in the AI or NLP literature. 
In this paper, we introduce the task of \textit{profiling}, inspired by theories and findings in social psychology about the potential of profiles for reasoning and information processing. % over long-tail instances.
%In this paper, we introduce the cognitive task of profiling as a computational counterpart to the theories and findings in social psychology and anthropology. 
We describe two generic state-of-the-art neural architectures that can be easily instantiated as profiling machines to generate expectations and applied to any kind of knowledge to fill gaps. We evaluate these methods against Wikidata and crowd expectations, and compare the results to gain insight in the nature of knowledge captured by various profiling methods. We make all code and data available to facilitate future research. 
\end{abstract}
%\todo{piek: the abstract is very ambitious}
\section{Introduction}
\label{sec:intro}

%{\bf EH: this version aims at AAAI} 
A key characteristic of intelligent behavior is the use of knowledge.  
The human mind is a powerful multifunctional knowledge storage and management system that performs generalization, type inference, anomaly detection, profiling, stereotyping, and other tasks.  
No computational system to date implements all these capabilities together. Traditional databases comprise one or more tables of (usually instantial) facts, coupled to a definitional metadata schema.  Declarative AI knowledge bases are essentially multigraph concept networks with explicitly named relations between nodes (each node being a cell in an equivalent database), allowing one to define filler constraints along the edges and perform inference via graph traversal.  KR typologies/ontologies encode and organize terminology and definitional information, but seldom include instance data.  %None of them can generalize from instances to concepts or detect anomalies natively.  \todo{piek: is this really true?}

Growing amounts of data and the advent of workable neural (deep) models raise the natural question: can one build models that simultaneously encode instance-level knowledge, generalize it to form `concepts', learn profiles, detect anomalies, provide preferences, etc.? How can one generalize and spot inconsistencies when, unlike in databases or Semantic Web, no explicit ontological schema can be assumed? Can one test such models in applications like natural language QA systems that require various semantic background knowledge exactly because stereotypical information is deliberately left out in people's communication?

%to overcome the omissions/inaccuracies typical of the way people express language?  %\todo{piek: i would not say that language itself is inaccurate. It can be as accurate as we want. Moreover it is people expressing themselves through language who are inaccurate or NLP not being able to handle underspecification that people resolve easily}%\todo{piek: In addition to inaccuracies in language, another use is to detect anomalies in databases. This may be more interesting even for AI audience}
%\todo{piek: i would tone down the last section. why not motivate this in a positive way? We need profiles to make predictions as needed knowledge is often lacking in applications. But there is hardly any research on this.}

In this paper we focus on an important but unaddressed component of this challenge: {\em profiling}.  We believe that a representation [learning] system should be able to absorb instances (exemplars) of some multifaceted concept (such as Person) and automatically generate expectations for the unspecified facets. These expectations should of course be conditioned on whatever (partial) information is provided for any test case, and automatically adjusted when any additional information is provided.  
Further, they should conform to human intuition and experience.  
%In our current work, we do not require probabilistic soundness, though the exploration of probability in such models is a fascinating topic that has received some attention.  

To the best of our knowledge, this quite natural task of profiling has not been addressed widely in the AI or NLP literature. A related challenge, knowledge base completion (KBC), has however been relatively popular: in it, the system is asked to add new concrete instance-level facts given other instantial information.
%provide specific instantial information for a single facet of a given instance, given values for its other facets. 
This can be seen as an `edge case' of profiling, but is fundamentally different in that it predicts specific values rather than generating human-like expectations over value classes. 
%A solid body of research in fields like social psychology and anthropology provide a cognitive justification and motivation for profiling instead of for KBC. \todo{piek: KBC sometimes makes no sense. What is the purpose of guessing somebodies precise age? It is more realistic and useful to predict value ranges}
%\todo{piek: just a comment. could you say that one is a subtask of the other? Maybe make the difference stronger by saying that one is more driven by ontological constraints and profiling more by historic facts?} 
%cannot answer questions for which it cannot infer specific values, though it can answer some questions that profiling simply cannot, such as ``what is the birth month of Person X?". 
%We hereby introduce the task of profiling over incomplete knowledge in the 
%Social psychology findings support our belief that %allow us to expect that
%We express our belief that 
%(cognitive) 
%Social psychology findings support our belief that AI systems which include an accessible profiling component can naturally fill knowledge gaps with assumed expectations. 
We see profiling as a common and potentially more helpful capability for NLU and QA, where knowing preferences or general expectations is often necessary in order to perform reasoning and to direct interpretation on partial and underspecified data. 
%it is often sufficient to know preferences or general expectations in order to direct 
%tasks.% such as entity disambiguation, parse tree and semantic structure attachment, coreference, (implicit) semantic role labeling, etc. \todo{piek: this point is too specific NLP for AAAI. keep it for TACL}
Profiling lies somewhere between the existing tasks of KBC in machine learning and missing value imputation in databases, seeking to fill {\em all} missing values (as in the field of databases), albeit with types, where the underlying entities and relations are analogous to those used in KBC.
%This perspective is not the first work in AI focusing on missing knowledge, but is the first that poses this problem from a cognitive perspective.  

We summarize the contributions of this paper as follows:
\begin{enumerate*}
\item We define the task of profiling over knowledge for AI machines, which is especially needed for contextual, long-tail instances, where knowledge is extremely scarce, yet much of what is unknown can be stereotyped. 
\item We pose this task as inspired by insights in social psychology, potentially providing another link between cognitive sciences and AI systems.
%to benefit from, but also contribute to, insights and theories in e.g. social psychology. 
\item We perform profiling on coarse-grained values to bridge the gap between factual data and human stereotypes.
\item We describe two generic state-of-the-art neural methods that can be easily instantiated into profiling machines for generation of expectations, and applied to any kind of knowledge to fill gaps. 
\item We evaluate these methods against Wikidata and crowd expectations, and compare the results to gain insight in the nature of knowledge captured by profiles.
\item We make all code and data available to facilitate future research. 
\end{enumerate*}

\section{Related Work}
\label{sec:relwork}

Several quite distinct research areas are relevant to profiling.  We briefly list some of the most relevant work.  

%From a technical perspective, the cognitive task of stereotyping over incomplete knowledge partially overlaps with the tasks of data imputation and KBC. Hence, in this Section we review some key research on these two tasks.

\textbf{Data imputation.} %Missing data in real-world databases restricts their usability in applications that rely on complete data, like reporting or vizualization tools. 
Data imputation refers to the procedure of filling in missing values in databases. 
In its simplest form, this procedure can be performed by mean imputation and hot-deck imputation~\cite{lakshminarayan1999imputation}. Model-based methods are often based on regression techniques or likelihood maximization~\cite{pearson2006problem}. \citeauthor{gautam2015data} combine a neural network with a genetic algorithm, while \citeauthor{aydilek2013hybrid} cluster missing values with existing, known values. 
These efforts focus on guessing numeric and interval-valued data, 
%similar to the task of guesstimation in the Linked Data community
which is a shared property with the related task of guesstimation
%addressed in the Linked Data community
~\cite{abourbih2010using}. In contrast, profiling aims to predict typical {\em classes} of values. Moreover, data imputation has no direct cognitive correlate regarding inferred expectations.% to infer. %Additionally, stereotyping focuses on predicting typical classes of values rather than specific (e.g. numeric) values.

\textbf{KBC.} 
%While only a relatively small proportion of values are missing in databases (the usual percentage in the evaluation datasets is between 5\% and 20\%), 
Compared to databases, the sparsity in real-world AI knowledge graphs is typically far greater: e.g., most facet values in both Freebase and Wikidata are missing~\cite{dong2014knowledge}. KBC adds new facts to knowledge bases/graphs based on existing ones. Two related tasks are link prediction (with the goal to predict the subject or object of a given triplet, usually within the top 10 results) and triplet completion (a binary task judging whether a given triplet is correct). 

In the past decade, KBC predominantly employed deep embedding-based approaches, which can be roughly divided into tensor factorization and neural network models~\cite{ji2016knowledge}. 
TransE~\cite{bordes2013translating} %TransH~\cite{bordes2014semantic}, 
% TransR~\cite{lin2015learning} and
and ITransF~\cite{xie2017interpretable} are examples 
%is an example 
of neural approaches that model the entities and relations in  embedding spaces, and use vector projections across planes to complete missing knowledge. Tensor factorization methods like \cite{%nickel2015review,
guu-miller-liang:2015:EMNLP} regard a knowledge graph as a three-way adjacency tensor. Most similar to our work, Neural Tensor Networks~\cite{socher2013reasoning} also: 
\begin{enumerate*}
\item aim to fill missing values to mimic people's knowledge; 
\item evaluate on structured relations about people; 
\item rely on embeddings to abstract from the actual people to profile information
\end{enumerate*}. 
%Unlike Socher et al., our task is to predict typical classes of values rather than 
%restrict the entire knowledge about each instance to mostly N attributes, while Socher et al.\ use all entity knowledge to predict just the 7 attributes of interest. Additionally, our methods do not need negative examples for learning.

% \citeauthor{socher2013reasoning} state that KBC can be used to answer questions like \textit{Does a dog have a tail?}. 
% \citeauthor{guu-miller-liang:2015:EMNLP} take a step further and propose a technique that captures graph paths, enabling one to, e.g. answer \textit{What is the nationality of Tad Lincoln?} based on the nationality of his father, Abraham Lincoln.
% The compositional method in \cite{neelakantan2015compositional} can reportedly handle paths that are unseen during training. 

%advance this idea towards compositionality, with a training method which captures graph paths based on the intermediate tensor products. This would answer \textit{What is the nationality of Tad Lincoln?} based on the nationality of his father, Abraham Lincoln. 

Universal Schema (US)~\cite{riedel2013relation} includes relations extracted from text automatically, which results in larger, but less reliable, initial set of relations and facts. US was designed for the needs of NLP tasks such as fine-grained entity typing~\cite{yao2013universal}.

KBC research, including that by \citeauthor{socher2013reasoning}, resembles the task of profiling in that it also operates on data about real-world instances and their typed relations. However, the goals of profiling differ from those of KBC: 
\begin{enumerate*}
\item to generate an expectation class for \textit{every facet} of a category/group, rather than suggesting missing facts;
\item to provide a \textit{typical distribution} (not a \textit{specific} value) for the attributes of a specific group;
\item to understand how these profiles shift when additional evidence is provided
\end{enumerate*}. 
These differences make profiling far more useful for reasoning over incomplete data in AI applications, and related to cognitive 
%Moreover, to our knowledge, KBC has never been related to cognitive 
work on stereotypes.
%analysed from a cognitive perspective. 

KnowledgeVault (KV)~\cite{dong2014knowledge} is a probabilistic knowledge base by Google, which fuses priors about each entity with evidence about it extracted from text. Despite using a different method, the priors in KV serve the same purpose as our profiles: they provide expectations for all unknown properties of an instance, learned from factual data on existing instances. Unfortunately, the code, the experiments, and the output of this work are not publicly available, thus preventing further analysis of these priors, the validity of factual data for profiling, and their relation to cognitive/cultural profiling as done by humans.

%You want to do the comparison against the crowd - interesting & points to the cognitive aspect + the difference between factual and cognitive/cultural profiling

%probabilistic knowledge base by Google. They a) start from Freebase, then b) generate prior expectations for all unknown properties of all instances, and then c) try to extract knowledge about the missing values from the web. The extraction in c. is compared and combined with the prior beliefs in b. - namely, the extracted facts  (in c) that are more intuitive (according to b) are accepted easier and have higher probabilities.

\textbf{Estimation of property values.} Past work in IR attempted to improve the accuracy of retrieval of long-tail entities by estimating their values based on observed head entities. %~\cite{reinanda2016document,farid2016lonlies}. %\citeauthor{reinanda2016document} propose a set of entity-independent features representing the aspects of informativeness, saliency and timeliness. 
Most similar to profiling, the method by \citeauthor{farid2016lonlies} estimates a property of a long-tail entity based on the community/ies it belongs to, assuming that each entity's property values are shared with others from the same community. Since the goal of this line of research is to improve the accuracy of retrieval, the generalization performed is rather ad-hoc, and the knowledge modeling and management aspects have not been investigated in depth. Moreover, the code, the experiments, and the results of this work are not publicly available for comparison or further investigation.

%propose an entity-independent method that can generalize over unseen or rarely seen entities. However, the goal of this research is not to estimate entity property values, but instead to aid the document filtering process in information retrieval.
%The method by \cite{farid2016lonlies} estimates a property of a long-tail entity based on the community/ies it belongs to. This approach is based on the assumptions that each entity belongs to a community and that its property values can be estimated through the property values of the similar entities from the same community. 
%``Community'' is extracted through co-occurrence with head (high-frequency) entities in the same documents, or through selecting the entities that have the same property relation with a same entity (e.g., all entities that work at CMU). 

\textbf{Social media analysis.} Community discovery and profiling in social media is a task that clusters the online users which belong to the same community, typically using embeddings representation~\cite{akbari2017leveraging}, without explicitly filling in/representing missing property values. 

Local models that infer a specific property (e.g., the tweet user's location) based on other known information,  such as her social network~\cite{jurgens2013s} or expressed content~\cite{mahmud2012tweet}, also address data sparsity. These models target specific facets and data types, thus they are not generalizable to others. Similarly to models in KBC, they lack cognitive support and fill specific instance values rather than typical ranges of values or expectations.
%Local generalization models with ideas similar to the ones on profiling

\textbf{Probabilistic models.} 
Prospect Theory~\cite{kahneman1979prospect} proposes human-inspired cognitive heuristics to improve decision making under uncertainty.
The Causal Calculus theory~\cite{pearl2009causality} allows one to model causal inter-facet dependencies in Bayesian networks. Due to its cognitive background and assumed inter-facet causality, profiling is a natural task for such established probabilistic theories to be applied and tested.

\textbf{Stereotypes.} 
Stereotype creation is enabled by profiling. The practical uses of stereotypes are vast and potentially ethically problematic. For example, \citeauthor{bolukbasi2016man} claim that embedding representations of people carry gender stereotypes; they show that the gender bias can be located in a low-dimensional space, and removed when desired.  We leave ethical and prescriptive considerations for other venues, and note simply that artificially removing certain kinds of profiling-relevant signals from the data makes embeddings far less useful for application tasks in NLP and IR when evaluated against actual human performance.% Eventually, profiles can be used to create system awareness of stereotypes and to act in an ethical way without `removing' this knowledge.

%We certainly agree with the authors that machine learning algorithms and applications of embeddings should be careful not to amplify gender biases more than they are already present. At the same time, we stress the importance of this type of implicit information, since the embeddings apparently capture stereotypes that are embedded in human communication; if this information is removed, the embeddings become much less useful for application tasks in NLP and IR. For instance, the authors point to an analogy of the system ``he to doctor as she to X''. The original embeddings answer with X=nurse, while after debiasing the answer is X=physician. While the latter is less likely to be offensive than the former, we also note that the former answer is reflecting human expectations much more accurately.

% \begin{itemize}
% \item knowledge completion/statistical relational learning/link prediction: Information extraction (e.g Socher and Chen, 13), Type extraction (Neelakantan et al., 2015), question answering over a KB graph (Guu and Liang 15) (e.g What languages are spoken by people living in Lisbon?)<see may 04 meeting notes>
% \item database missing values amputation ((Berkan and Arslan, 2011), (Ssali and Marwala, 2007), (Mohamed et al., 2007))
% \end{itemize}

\section{Profiles}
\label{sec:profiles}

\subsection{Intuitions}
People have no trouble making and using profiles/defaults: 

\indent \textit{P1 is male and his native language is Inuktitut. What are his citizenship, political party, and religion? Would knowing that he was born in the 19th century change one's guesses?}

%\textit{P2 is a female Russian politician born in Moscow. Can you guess her political party and cause of death? What would be her cause of death in case she is a member of the party ``United Russia''?}
%\qizhe{Is the example making people a little uncomfortable?}

\textit{P2 is a member of the American Senate.  Where did he get his degree, and what is his native language?}

\textit{P3 is an army general based in London.  What is P3's stereotypical gender and nationality?  If P3 gets an award ``Dame Commander of the Order of the British Empire'', which expectations would change?}

Presumable answers to the above questions are as follows. P1 is a citizen of Canada, votes for the Conservative Party of Canada, and is Catholic. However, P1's 19th century equivalent belongs to a different party. P2 speaks English as main language and graduated at Harvard or Yale University. Finally, the initial expectation on P3 of a male Englishman switches to a female after the award evidence is revealed.%\todo{piek: this is simply an inference that Dame implies female}

Most of us would agree with the suggested answers. Why/how!? What is it about the Iniktitut language that associates it to the Conservative Party?  Why is the expectation about the same person in different time periods different?  Why does the sole change of political party change the expectation of one's work position or cause of death?  Despite the vast set of possibilities, these kinds of knowledge gaps are easily filled by people based on knowledge about associations among facet values, and give rise to a continuously evolving and changing collection of cognitive expectations and stereotypical profiles. 
%In human knowledge as well as communication, profiles are implicit but crucial to replace the missing information.  
Further, people assume that they are entitled to fill knowledge gaps with their expectations unless contradictory evidence is explicitly presented; if the truth differs from the profile, pragmatics requires (as reflected in Grice's Maxim of Quantity) that this truth will be explicitly stated.\footnote{We would typically only state the native language of an American citizen if it is not English, unless that could be inferred from his/her name or other extant information.}

% Following the pragmatic rules of human communication, we assume that certain knowledge is being possessed by our reader, and for reasons of interestingness and brevity, we leave this information out. The target audience has no problem complementing this knowledge with the assumed prior expectations. 

AI systems such as NLU and QA engines require such human-like expectations. As this information tends to be deliberately absent from (or implicit in) human communication, information harvesting systems cannot simply distill it directly from available material. Knowledge missing during AI system processing is traditionally injected from knowledge bases, which attempt to mimic the world knowledge possessed and applied by humans.
However, current knowledge bases are notoriously sparse~\cite{dong2014knowledge}, and do not contain implicit expectations nor provide inference mechanisms that can generate such expectations. AI systems face an apparently insurmountable obstacle, most notable for long-tail instances with little to no accessible facts.\footnote{e.g.~\cite{esquivel2017long} reports that around 50\% of the people mentioned in news documents are not represented in Wikipedia.}%  leaving all other information either implicit or assumed to be unimportant.

% Knowledge missing during AI system processing is traditionally injected from knowledge bases, which attempt to mimic the world knowledge possessed and applied by humans.  Knowledge bases have been proven useful in many applications in NLP and IR. However, most knowledge bases are notoriously sparse.  For example, Wikidata records native language information for only 1.5\% and birthplace information for 36\% of its Persons.  
% Unfortunately, current knowledge bases neither contain implicit expectations nor provide inference mechanisms that can generate such expectations.   
% But as we show later, much of what is ``unknown'' can quite easily be obtained from stereotypical profiles.  

% Without a general method to generate expectations from existing incomplete knowledge, knowledge bases and their applications are deprived of this vast implicit knowledge. The significance of knowledge profiling notably increases once we switch from popular instances toward the long tail of instances, relevant only within specific micro-contexts. Such tail instances will usually not appear in a dataset,\footnote{e.g.~\cite{esquivel2017long} reports that around 50\% of the people mentioned in news documents are not represented in Wikipedia.} or may be described with only a few key facts, leaving other facts either implicit or assumed to be unimportant.

\subsection{Aspects of Profiles}

Following~\cite{ashmore1981conceptual} we define a profile as a set of beliefs about the attributes of members of some group. A stereotypical profile is a type of \textit{schema}, an organized knowledge structure that is built from experience and carries predictive relations, thus providing a theory about how the social world operates~\cite{kunda1999social}. Stereotypical profiles shared by a culture/sample are named {\em consensual}.%\footnote{Here we focus on the cognitive function, content, and accuracy of consensual profiles. Other aspects: their social role, relation to discrimination and prejudice, etc. fall out of the scope of this paper and should be considered in future research.}

{\bf Functions.} As a \textit{fast} cognitive process, profiling gives basis for acting in uncertain/unforeseen circumstances~\cite{dijker1996stereotyping}.
Profiles are ``shortcuts to thinking'', that provide people with rich and distinctive information about unseen individuals.
Moreover, they are \textit{efficient}, thus preserving our cognitive resources for more pressing concerns.
%and \textit{resource-preserving}, %llowing us to redirect our cognitive energy on more pressing concerns.

% profiling provides two main qualities~\cite{dijker1996stereotyping}: \begin{enumerate*}
% \item being \textit{fast}, it gives people basis for acting in uncertain/unforeseen circumstances. Profiles are ``shortcuts to thinking'', that provide us with rich and distinctive information about unseen individuals.
% \item being \textit{efficient} and \textit{resource-preserving}, they allow us to redirect our cognitive energy on more pressing concerns
% \end{enumerate*}. %\citeauthor{bodenhausen1988stereotypic} notes the potential downside of these functions: people occasionally ignore  information that is relevant but inconsistent with the stereotypical profile.

%They can be {\em consensual} (stereotypes shared by a particular culture or sample) or {\em personal} (individual's beliefs). 

{\bf Accuracy.} Profile accuracy reflects the extent to which beliefs about groups correspond to the actual characteristics of those groups~\cite{jussim2015stereotype}. Consensual ones have been empirically shown to be highly accurate, especially the demographic (race, gender, ethnicity, etc.)\ and other societal profiles (like occupations or education), and somewhat less the political ones~\cite{jussim2015stereotype}. This effect has been called the ``wisdom of the crowds''~\cite{surowiecki2004wisdom}.

This high accuracy does not mean that profiles will correctly describe any group individual; they are a statistical phenomenon.\footnote{This is a key difference with KBC. KBC looks for the \textit{correct} facet value for an individual, while profiling for the \textit{typical} one.} In that sense, the findings by \citeauthor{jussim2015stereotype} that most profiles are justified empirically are of great importance for AI machines: it means that they can be reliably inferred from individual facts, which (unlike many profiles themselves) are readily available.%\todo{piek: makes me think that profiling can also detect gradual changes in expectations (learn by observing) which is not dealt with in KBC.}

\textbf{Granularity.} Profiles exist at various levels of specificity for facets and their combinations. A profile of 20th century French people differs from a profile of 20th century people in general, with more specificity in what kind of food they eat and what movies they watch, or from the profile of French people across all ages. Added information usually causes the initial expectations to change (``shift''), gradually narrowing the denotation group in a transition toward ultimately an individual. The shift may be to a more accurate profile (what in \cite{stone1957two} is called an ``accurately shifted item''), or the opposite (``inaccurately shifted item'').  %According to these authors, people tend to be good either with sticking with profiles or departing from them, but typically not both. 

%{\bf This is discussed in related work -- EH} 
%\textbf{Reason, Ethics, Opportunity.} As discussed by \citeauthor{jussim2015stereotype}, the value of a profile should be judged solely on the basis of empirical experimentation.  Stereotyping, closely related to profiling, is rather controversial when it comes to ethics, since it might be used for prejudice and discrimination. In this direction, \citeauthor{bolukbasi2016man} illustrate a method to remove the gender bias from embeddings. 

%Profiles play a central role in human knowledge and communication, and serve as expectations allowing humans to fill knowledge gaps. The fact that stereotypes largely fit the instance data is a vast opportunity for AI machines to complete knowledge in a human-like manner.  

\subsection{Formal Definition of a Profile}

Given a finite set of $n$ facets, each \textbf{group} $g$ is defined through a set of $k$ known attribute values, namely: $$g(X, Y, K)=\{\bigcup_{i=1}^{k} (x_i, y_{ij})| x_i\in X \land y_{ij} \in Y_i\}$$
$X=x_1,..,x_n$ is the set of all facets, $Y_i=y_{i1},..$ is the set of possible values for the attribute $x_i$, as found in the background knowledge $K$. Concrete groups can be easily instantiated from this definition: \{(`religion', `Buddhism')\} is a group of all Buddhists, \{(`religion', `Buddhism'), (`citizenship', `USA')\} is a group of all American Buddhists, etc.%whereas \{(`religion', `Buddhism'), (`sex or gender', `female')\} defines the group of all female Buddhists.

%Conceptually, we defined a \textbf{profile} as a set of beliefs about the unknown attributes of members of a group. 
The \textbf{profile} $pr$ for a group $g={(x_1, y_{1j}), ..., (x_k, y_{kj})}$ can be defined as a distribution of expected values (probabilities) for each of the remaining $(n-k)$ properties, namely:
$$pr(g, X, Y, K)=g(X,Y,K) \cup \{(x_{k+1}, d_{k+1}),...,(x_n, d_n)\}$$
where $d_{k+1},...,d_n$ are distributions of expected values for the properties $x_{k+1},...,x_n$ given the known property values in the group $g$. For a given group, a profile is chosen to be optimal when its property-value pairs have the highest probability given the background knowledge used for training.

\section{Profiling Methods and Implementation}
\label{sec:approach}

In this section we describe two neural architectures for computing profiles at large scale, and baselines for comparison.

\subsection{Autoencoder (AE)\footnote{Demo at %\url{http://stereotypes.fii800.lod.labs.vu.nl/}
 $<anonymized link>$}
}
\label{ssec:ae_model}

%\todo{write more about the input x}

%\todo{give example of the attributes and what's the corresponding x like}

An autoencoder is a neural network that is trained to copy its input to its output~\cite{Goodfellow-et-al-2016}. Our AE contains a single densely connected hidden layer. Its input \textbf{\textit{x}} consists of $n$ discrete facets $x_1, ..., x_n$, where each $x_i$ is encoded with an embedding of size $N_e$. 
For example, if the input is the entity Angela Merkel, we concatenate $n$ embeddings for its $n$ individual features (nationality: German, political party: CDU, etc.).
The total size of the embedding input layer is then $|\textit{\textbf{x}}|=n*N_e$.

We denote the corresponding output for an input sequence \textbf{\textit{x}} with $\textit{\textbf{z}}=\textit{g(f(\textbf{x}))}$, where $f$ is the encoding and $g$ is the decoding function. The output layer of the AE assigns probabilities to the entire vocabulary for each of the $n$ features. 
%Since the vocabulary size usually varies across the features, 
The size of the output layer is a sum over the variable vocabulary sizes of the individual inputs $v_i$: $ |\textit{\textbf{z}}| = \sum_{i=1}^{n} v_i$.

The AE aims to maximize the probability of the correct class for each feature given inputs \textbf{\textit{x}}, i.e., it is trained to minimize the cross-entropy loss $L$ that measures the discrepancy between the output and the input sequence: $$\textit{L(\textbf{x}, \textbf{z})}=-\sum_{i=1}^{n}[x_ilogz_i+(1-x_i)log(1-z_i)]$$ 

%\todo{when non-existing value(""), we mask its value while evaluating. When a value out of the range of accepted classes (UNK), we give it a special value (1).}
Due to the sparse input, it is crucial that AE can handle missing values. We aid this in two ways: 
\begin{enumerate*} 
\item we deterministically mask all non-existing input values;
\item we apply a probabilistic dropout on the input layer, i.e., in each iteration we randomly remove a subset of the inputs (any existing input is dropped with a probability \textit{p}) and train the model to still predict these correctly. Although we apply the dropout method to the input layer rather than the hidden layer, we share the motivation with \citeauthor{srivastava2014dropout} to make the model robust and able to capture subtle dependencies between the inputs.  Such dropout helps the AE reconstruct missing data. 
%We adapt this idea with a notable difference of applying a dropout regularization to elements of the input layer rather than the hidden layer. Intuitively, this design decision would encourage the network to capture more subtle and less obvious dependencies between the inputs.
\end{enumerate*}

%\todo{give this more space: usual dropout drops hidden layer, but here we mask some input - traditional dropout is continuous, here it is discrete}

%\todo{not training jointly, but instead training the AE
%: the embeddings are part of the AE.}

%\todo{upsampling, adam, cross entropy, put in implementation details}

%\todo{feed-forward NN equations}

%During this process, we jointly train the weights of the autoencoder and the embeddings of our input values. 

%\todo{(later) if the community is less modern AI, specify the embeddings further}

%\todo{(later) maybe it makes sense to have a schema/graph}

\subsection{Embedding-based Neural Predictor (EMB)}
\label{ssec:em_model}

In our second architecture each input is a single embedding vector $\textbf{e}$ rather than a concatenation of $n$ facet-embeddings. For example, the input for the entity Angela Merkel is its fully-trained entity embedding. The size of the input is the size of that embedding: $|\textit{\textbf{e}}|=N_e$. In the current version of EMB, we use pretrained embeddings as inputs and we fix their values (these were trained in a previous project). Future work can investigate the benefits of further training/tuning, or even training from scratch for cases where pre-trained embeddings are not available.%.\footnote{As pre-trained embeddings are often not available or easy/quick to create, sometimes such training is unavoidable.}

%For example, the input for the entity Angela Merkel would be its entire entity embedding. 
%In our current approach we have fixed the pretrained embeddings, but 

Like the AE, EMB has one densely connected hidden layer. For an input \textbf{\textit{x}} and its embedding representation \textbf{\textit{e}}, the corresponding output is $\textit{\textbf{z}}=\textit{g(h(\textbf{e}))}$. The output layer of the embedding-based predictor has the same format as the one of the AE, and the same cross-entropy loss function $\textit{L(\textbf{x},\textbf{z})}$.
%$$\textit{L(\textbf{x}, g(h(\textbf{e}))}=-\sum_{i=1}^{n}[x_ilogz_i+(1-x_i)log(1-z_i)]$$

%\qizhe{Needs to mention how the entity embedding is obtained}

\subsection{Baselines}
We evaluate the methods against two baselines:

\textbf{Most frequent value baseline (MFV)} chooses the most frequent value in the training data for each attribute, e.g., since $14.26\%$ of all politicians are based in the USA, MFV's accuracy for profiling politicians' citizenship is $14.26\%$. This baseline indicates for which facets and to which extent our methods can learn dependencies that transcend MFV. 
%help us understand for which attributes our methods learn complex dependencies, as opposed to those for which they mainly get biased to the most common value. 
%For example, 13\% of all politicians are based in USA: thus the top-1 accuracy of the MFV baseline on predicting citizenship for politicians is 13\%.

\textbf{Naive Bayes classifier (NB)} applies Bayes' theorem with strong independence assumptions between the features. We represent the inputs for this classifier as one-hot vectors. Naive Bayes classifiers consider the individual contribution of each input to an output class probability.  However, the independence assumption prevents it from adequately handling complex inter-feature correlations.% (e.g. one's stereotypical place of death might depend on a combination of: place of birth, job position, cause of death, nationality, gender, etc.).

\subsection{Model Implementation Details}

Both neural models use a single dense hidden layer with 128 neurons. For the AE model, we pick an attribute embedding size of $N_e=30$. These vectors are initialized randomly and trained as part of the network. We set the dropout probability to $p = 0.5$. The inputs to the EMB are 1000-dimensional vectors that were previously trained on  Freebase.\footnote{These vectors are available at \url{code.google.com/archive/p/word2vec/}. They correspond and can be mapped to only a subset of all Wikidata entities.} 

Both models were implemented in Theano~\cite{bergstra2010theano}. We used the ADAM \cite{kingma2014adam} optimization algorithm. We train for a maximum of 100 epochs with early stopping after 10 consecutive no-improvement iterations, to select the best model on a held-out validation data.%\footnote{We stop training after 10 consecutive evaluations with no improvement.} 
We fix the batch size to 64. When an attribute has no value in an entire minibatch, we apply oversampling: we randomly pick an exemplar that has a value for that attribute from another minibatch and append it to the current one.

\begin{table*}[!ht]
\small
\centering
\begin{tabular}{r | c c c c | c c c c | c c c c }
  & \multicolumn{4}{c}{\textit{PERSON}}  & \multicolumn{4}{c}{\textit{POLITICIAN}} & \multicolumn{4}{c}{\textit{ACTOR}}     \\
\textbf{attribute}  & $\mathbf{n_{ex}}$ & $\mathbf{v_i}$ & $\mathbf{H_i}$ & $\mathbf{H_i\prime}$  & $\mathbf{n_{ex}}$ & $\mathbf{v_i}$ & $\mathbf{H_i}$ & $\mathbf{H_i\prime}$ & $\mathbf{n_{ex}}$ & $\mathbf{v_i}$ & $\mathbf{H_i}$ & $\mathbf{H_i\prime}$ \\
\toprule 
educated at & 273,096 & 3,000 & 9.28 & 0.80 & 22,461 & 3,000 & 9.73 & 0.84 & 5,047 & 883 & 7.56 & 0.77 \\ \midrule
sex or gender & 2,403,980 & 11 & 0.64 & 0.18 & 168,758 & 5 & 0.50 & 0.25 & 75,980 & 5 & 1.00 & 0.50  \\ \midrule
citizenship & 1,546,757 & 995 & 5.28 & 0.53 & 152,131 & 335 & 5.07 & 0.61 & 57,570 & 187 & 5.12 & 0.68  \\ \midrule
native language & 41,760 & 141 & 1.70 & 0.24	& 16,818 & 33 & 1.08 & 0.21 & 4,273 & 29 & 0.41 & 0.08  \\ \midrule
position held & 177,302 & 3,000 & 7.44 & 0.64 & 101,766 & 1,701 & 7.08 & 0.66 & 244 & 25 & 0.96 & 0.21  \\ \midrule
award received  & 154,275 & 3,000 & 7.97 & 0.69 & 10,588 & 546 & 6.82 & 0.75 & 2,880 & 297 & 6.60 & 0.80 \\ \midrule
religion  & 32,311 & 341 & 3.24 & 0.38 & 2,414 & 127 & 3.99 & 0.58 & 164 & 24 & 2.47 & 0.56 \\ \midrule
%participant of  & 104,058 & 919 & 2,863 & 53 & 34 & 28 & 10 & 11 \\ \midrule
political party  & 158,105 & 3,000 & 7.28 & 0.63 & 82,617  & 2,456 & 7.26 & 0.64 & 232 & 53 & 3.23 & 0.58 \\ \midrule
work location & 68,602 & 1,989 & 6.25 & 0.57 & 30,320 & 272 & 5.07 & 0.63 & 116 & 41 & 3.99 & 0.74  \\ \midrule
place of death & 350,720 & 3,000 & 7.93 & 0.68 & 29,071 & 3,000 & 8.39 & 0.73 & 9,377 & 2,169 & 8.33 & 0.75 \\ \midrule
place of birth & 927,089 & 3,000 & 7.64 & 0.66 & 59,627 & 3,000 & 7.27 & 0.63 & 39,694 & 3,000 & 8.55 & 0.74 \\ \midrule
cause of death & 21,926 & 499 & 5.35 & 0.60 & 1,408 & 115 & 4.75 & 0.69 & 1,039 & 82 & 4.22 & 0.66 \\ \midrule
lifespan range & 922,634 & 55 & 1.89 & 0.33 & 79,346 & 39 & 1.68 & 0.32 & 19,055 & 11 & 1.77 & 0.49  \\ \midrule
century of birth & 1,975,197 & 43 & 1.36 & 0.25 & 140,087 & 22 & 1.48 & 0.33 & 61,506 & 11 & 0.56 & 0.16  \\ 
\bottomrule
\end{tabular}
\caption{Numbers of examples ($n_{ex}$), categories ($v_i$), and entropy ($H_i$ and $H_i\prime$) per facet of People in our training data. We limit $v_i$ to 3,000 to restrict the complexity of the value space, but also to mimic the simplification aspect of cognitive profiling.}
\label{tab:stats}
\end{table*}

\section{Evaluation on Wikidata Instances}
\label{sec:setup}

\subsection{Data}

No existing dataset is directly suitable to evaluate profiling.  We therefore chose People, since data is plentiful, people are multifaceted, and it is easy to spot problematic generalizations.  We defined three typed datasets: people, politicians, and actors, each with the same stereotypical facets, such as nationality, religion, and political party, that largely correspond to some facets central in social psychology research. We created data tables by extracting facets of people from Wikidata. Table~\ref{tab:stats} lists all attributes, each with its number of distinct categories $v_i$, total non-empty values ($n_{ex}$), and entropy values ($H_i$ and $H_i\prime$) on the training data. 
%\todo{piek: explain which columns and column labels mark this}

%Table~\ref{tab:stats} also contains training data statistics on the number of distinct categories and total non-empty values per facet.% for the four full datasets we evaluate on. 

%We train our models and evaluate their performance on data about people found in Wikidata. We build one classifier per type/profession. In our current experiments, we have evaluated on four types: People, Politicians, Actors and Lawyers, but expanding this to any type is only a matter of extracting the data and training the neural architectures.

%We pick 15 general attributes characteristic for people and reuse them for all datasets: educated at, sex or gender, country of citizenship, native language, position held, award received, religion, participant of, political party, work location, place of death, place of birth, cause of death, lifespan range, and century of birth.

The goal is to dynamically generate expectations for the same set of 14 facets in each dataset.  We evaluate on multiple datasets to test the sensitivity of our models to the number of examples and categories. The largest dataset describes 3.2 million people, followed by the data on politicians and actors, smaller by an order of magnitude.  As pretrained embeddings are only available for a subset of all people in Wikidata (cf. \textit{Model Implementation Details}), we cannot evaluate EMB directly on these sets. Hence, to facilitate a fair comparison of both our models on the same data, we also define smaller data portions for which we have pretrained embeddings.
%\footnote{The amount of lawyers with embeddings is only a few hundreds, so we do not evaluate on the restricted lawyers data.} 
%For all datasets, 
We randomly split each of the datasets into training, development, and test sets at 80-10-10 ratio.

\subsection{Quantification of the Data Space}

We quantify aspects of profiling through the set of possible outcomes and its relation to the distribution of values.

The total size of the data value space is 
  $d_{size} = \prod_{i=1}^{n} {v_i}$, 
where $n$ is the number of attributes and $v_i$ is the size of the category vocabulary for an attribute $x_i$ (e.g. $v_i=|$\{Swiss, Dutch\ldots \}$|$ for $x_i=nationality$). We define the \textbf{average training density} as the ratio of the total data value size to the overall number of training examples $n_{ex}$: 
  $d_{avg-d} = d_{size} / n_{ex}$. As an illustration, we note that the full dataset on People has $d_{size}=10^{39}$ and $d_{avg-d}=10^{32}$.

For the $i$-th attribute $x_i$, the entropy \textit{$H_i$} of its values is their `dispersion' across its 
%category value set (which has 
$v_i$ different categories. The entropy for each category \textit{j} of $x_i$ is computed as $-p_{i,j} log p_{i,j}$, where 
$ p_{i,j} = n_{ex}(i,j) / n_{ex}(i) $. The \textbf{entropy} of $x_i$ is then a sum of the individual category entropies: $ H_i = -\sum_{j=1}^{v_i} p_{i,j} log p_{i,j} $, whereas its \textbf{normalized entropy} is limited to $[0,1]$:
$ H_i\prime = H_i/log_2(n_{ex}(i)) $.
 Entropy is a measure of informativeness: when $H_i\prime = 0$ there is only one value for $x_i$; when all values are equally spread the entropy is maximal, $H_i\prime=1$ (with no MFV).  

Of course, we do not know the true distribution but only that of the sparse input data.  Here we assume our sample is unbiased. Table~\ref{tab:stats} shows that, e.g., {\em educated at} consistently has less instance values and a `flatter' value distribution (= higher $H_i\prime$) than {\em sex or gender}, where the category \textit{male} is dominant on any dataset, except for actors. The entropy and the categories size together can be seen as an indicator for the relevance of a facet for a dataset, e.g., $H_i\prime$ and $v_i$ of \textit{position held} are notably the lowest for actors. %, even compared to the smaller data on lawyers. 
We expect MFV to already perform well on facets with low entropy, whereas higher entropy to require more complex dependencies.

\begin{table*}[!ht]
\centering
\small
\begin{tabular}{r | c c c c | c c c c | c c c c }
 & \multicolumn{4}{c}{\textit{PERSON}}  & \multicolumn{4}{c}{\textit{POLITICIAN}} & \multicolumn{4}{c}{\textit{ACTOR}} \\
\textbf{attribute}    & \textbf{MFV} & \textbf{NB} & \textbf{AE} & \textbf{EMB} & \textbf{MFV} & \textbf{NB} & \textbf{AE} & \textbf{EMB} & \textbf{MFV} & \textbf{NB} & \textbf{AE} & \textbf{EMB}    \\
\toprule 
educated at	& 4.41  & 9.22 & 13.20 & \textbf{22.45} & 2.57 & 6.88 & \textbf{13.14} & 9.47 & 11.32 & 15.09 & 3.77 & \textbf{46.43} \\ \midrule
sex or gender & 82.61 & 81.76 & 82.37 & \textbf{95.83} & 85.15 & 84.10 & 83.23 & \textbf{94.79} & 49.71 & 57.97 & 55.20 & \textbf{89.06} \\ \midrule
citizenship  & 29.10 & 57.36 & 66.49 & \textbf{78.49} & 18.27 & 46.75 & 72.94 & \textbf{77.96} & 17.99 & 39.94 & 60.77 & \textbf{65.05} \\ \midrule
native language	& 44.70 & 69.44 & \textbf{87.63} & 33.33 & 46.67 & 88.89 & \textbf{93.33} & 83.33 & \textbf{95.00} & \textbf{95.00} & \textbf{95.00} & 91.67 \\ \midrule
position held 	& 8.44 & 32.92 & \textbf{45.66} & 21.43 & 15.47 & 28.93 & \textbf{45.03} & 41.18 & 50.00 & 50.00 & 50.00 & \textbf{100.0} \\ \midrule
award received 	& 4.98 & 15.95 & 21.56 & \textbf{37.50} & 3.85 & 10.58 & 18.27 & \textbf{26.09} & 14.29 & 14.29 & 23.81 & \textbf{42.86} \\ \midrule
religion 	& 27.52 & 40.83  & 45.48 & \textbf{71.43} & 27.08 & 42.71 & 52.08 & \textbf{56.52} & 40.00 & 40.00 & 60.00 & \textbf{66.67} \\ \midrule
political party	& 13.18 & 29.67 & 42.08 & \textbf{47.06} & 9.41 & 22.78 & 34.28 & \textbf{37.59} & \textbf{50.00} & \textbf{50.00} & \textbf{50.00} & 0.0 \\ \midrule
work location & 22.47 & 57.18 & \textbf{64.49} & 60.00 & 22.22 & 69.90 & \textbf{83.09} & 75.00 & 0.00 & 0.00 & 0.00 & 0.00 \\ \midrule
place of death	& 4.09 & 25.09 & 28.20 & \textbf{36.84} & 2.81 & 8.03 & 17.27 & \textbf{25.81} & 9.78 & 17.58 & 18.48 & \textbf{33.93} \\ \midrule
place of birth	& 2.85 & 33.01 & 32.07 & \textbf{49.04} & 1.88 & \textbf{54.62} & 23.59 & 52.21 & 5.31 & 11.28 & 16.87 & \textbf{36.21} \\ \midrule
cause of death	& 23.80 & 24.13 & \textbf{24.24} & 15.38 & 32.76 & 37.93 & 24.14 & \textbf{71.43} & 33.33 & 33.33 & 20.00 & \textbf{45.00} \\ \midrule
lifespan range	& 41.76 & \textbf{43.56} & 41.69 & 42.03 & 41.30 & 40.68 & 38.51 & \textbf{48.75} & 36.73 & 39.17 & \textbf{45.92} & 43.33 \\ \midrule
century of birth & 82.04 & 85.45 & 84.94 & \textbf{89.53} & 76.13 & 80.13 & 83.14 & \textbf{85.79} & \textbf{93.62} & 93.60 & 89.56 & 92.67 \\ 
\bottomrule
\end{tabular}
\caption{Top-1 accuracies for the both neural methods and the two baselines on the smaller datasets. For each dataset-facet pair, we emphasize the best result. Our neural methods, especially EMB, outperform the baselines. Entropy and vocabulary sizes can partially explain deltas in accuracies on individual facets.}
\label{tab:results_restricted}
\end{table*}
%\todo{piek: can you mark best and important results in bold in the table? Also add more explanation to the caption. People should more or less understand the table without reading the text.}

\begin{figure*}
\centering
\includegraphics[scale=0.5]{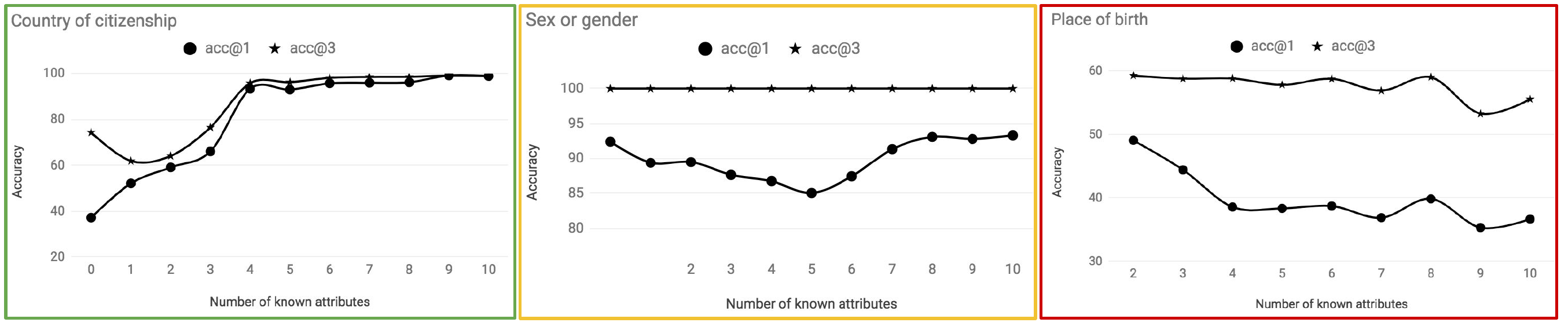}
\caption{Dependency of the accuracy of profiling politicians on the number of known facets: a positive correlation for \textit{country of citizenship} (left Figure), no correlation for \textit{sex or gender} (center), and a slightly negative one for \textit{place of birth} (right).}
\label{fig:known_attr}
\end{figure*}

\subsection{Results}
\label{ssec:results}

We evaluate by measuring the correctness of predicted (i.e., top-scoring) attribute values against their (not provided) true values, evaluated only on exemplars that were not included in the training data.  

Table~\ref{tab:results_restricted} provides the results of our methods and baselines on the three smaller datasets that contain embeddings (the full datasets yielded similar results for MFV, NB, and AE). We observe that AE and EMB outperform the baselines on almost all cases. As expected, we see lower (or no) improvement over the baselines for cases with low entropy (e.g., \textit{sex or gender} and \textit{lifespan range}) compared to attributes with high entropy (e.g., \textit{award received}). %Curiously, AE performs equally as the baselines for \textit{position held} only on the actors dataset, where also the entropy for this facet is the lowest.
We also note that the accuracy of profiling per facet correlates inversely with its vocabulary size $v_{i}$.

%Table~\ref{tab:results_full} shows that our AE outperforms the baselines on the four full datasets.  

%Table~\ref{tab:results_restricted} provides the results of our methods and baselines on the three smaller datasets that contain embeddings. 
The superiority of the neural models over the baselines means that capturing complex inter-facet dependencies improves the profiling ability of machines. Moreover, although the two neural methods perform comparably on average, there are differences between their accuracy on individual facets (e.g., compare \textit{award received} and \textit{native language} on any dataset). This might be due to the main architectural difference between these two methods: EMB's input embedding contains much more information (both signal and noise) than what is captured by the 14 facets in the AE.

% Table~\ref{tab:results_full} presents the results of our AE against the baselines on the four full datasets, and Table~\ref{tab:results_restricted} shows the results of our methods and baselines on the three smaller datasets that contain embeddings. First, these results show the superiority of the neural models over the two baselines, which means that capturing complex inter-facet dependencies improves the profiling ability of machines. 
% Second, the improvement over the baselines is typically lower on attributes with low entropy (e.g. \textit{sex or gender} and \textit{lifespan range}) compared to attributes with high entropy (e.g. \textit{award received}), which is largely expected. Curiously, the improvement for \textit{sex or gender} is exceptionally large for the actors datasets, on which the entropy for this attribute is by far the largest.
% Third, although the two neural methods perform comparably on average, we observe differences between their accuracy on individual facets (e.g., compare \textit{award received} and \textit{native language} on any dataset). This might be due to the main architectural difference between these two methods: EMB's input embedding contains much more information (both signal and noise) than what is captured by the 14 facets in the AE.
% Finally, it is intuitive that the accuracy of profiling per attribute correlates inversely with its vocabulary size $v_{i}$.

How does a profile improve (or not) with increasing input?  
We analyze both top-1 and top-3 accuracies of AE for predicting a facet value against the number of other known facets provided at test time. Figure~\ref{fig:known_attr} shows examples of all three possible correlations (positive, negative, and none) for politicians.
%plots three 
%Based on the outcome, we classify the attributes into: a) attributes for which knowing more about the specific instance improves the stereotyping accuracy (Figure~\ref{fig:known_attr}-left); b) attributes for which more instance-specific knowledge neither helps, nor harms the accuracy of stereotyping (Figure~\ref{fig:known_attr}-middle); c) attributes for which more instance-level knowledge leads to a decline in the stereotyping accuracy (Figure~\ref{fig:known_attr}-right). 
These findings are in line with conclusions from social psychology (cf. \textit{Profiles}): %\todo{piek: I am not sure that what you do matches their prediction. You correlate with the number of known attributes but that is not the same as the number of known instances is it? Please clarify.} 
knowing more facets of an instance might trigger a shift of the original profile, and it might be correct or incorrect, as defined in~\cite{stone1957two}. Generally, we expect that attributes with large $v_i$, like {\em place of birth}, will suffer as input exemplars become more specified and granularity becomes tighter, while facets with small $v_i$ would benefit from additional input. Figure~\ref{fig:known_attr} follows that reasoning, except for \textit{sex or gender}, whose behavior is additionally influenced by low entropy (0.25) and strong frequency bias to the \textit{male} class.

%So it is a little surprising that {\em religion} suffers from more information: subsequent work should investigate the source of this result.

%the source of this phenomenon should be investigated in .

% problems of evaluation metrics: to be discussed later 

%We discussed in the Section \textit{Data} that the data we evaluate on has the drawback of not containing profiles (what is \textit{typical}) but rather instance-level knowledge (what is \textit{correct}). This data is a fine approximation because we expect that most instances fit their profiles (since most profiles have been proven to be empirically justified). Still, the evaluation can be improved, either by creating a new dataset that directly evaluates profiles; 

%The underlying data could be sampled from various instances found in Wikidata in accordance to research in social psychology, while the annotation task should be suitable for crowdworkers. Another option is to evaluate the 

%or by an extrinsic evaluation on an application task in NLP or IR. We welcome ideas by AI researchers on how to optimally evaluate profiles. 

%\input{results.tex}

\section{Human Evaluation}

%\subsection{Data}
%\todo{piek: motivate why this experiment is important}
In order to collaborate with humans, or understand human language and behavior, both humans and AI machines are required to fill knowledge gaps with assumed world expectations (cf. \textit{Introduction}). 
Given that in most AI applications information is created for humans, a profiler has to be able to mimic human expectations. 
We thus compare our neural profiles to profiles generated by crowd workers.

\subsection{Data}

We evaluate on 10 well-understood facets describing American citizens.
%The final set of facets we used, ordered by increasing entropy, can be found in Table \ref{tab:human_results}. 
For each facet, we generated a list of 10 most frequent values among American citizens in Wikidata, and postprocessed them to improve their comprehensibility.
%compare the profiles generated by our AE system to human judgments. For this purpose, we chose 10 well-understood facets.  
%In a postprocessing step, we merged certain values within the top-10 list (e.g., ``Hollywood'' with ``Los Angeles''), and we removed too infrequent (e.g., non-binary genres) or non-sensible values (e.g., ``USA'').  
% We seek to study: 
% \begin{enumerate*}
% \item the relation between entropy and accuracy of profile generation
% \item the effect of adding/removing evidence on the prediction of the remaining attributes
% \end{enumerate*}.
% For this purpose, we collected Wikidata instances with known values for at least the 6 facets with either lowest or highest entropy. From each instance, we then created several less complete instances by removing one attribute at a time. For example, given that we know the 7 attributes with lowest entropy, we create an input instance with the top 7, the top 6, ..., and the top 3 attributes. Following this procedure, we created 191 rows with known low entropy attributes, and 114 rows where the highest entropy facets were known, resulting in 305 rows in total. 
We collected 15 judgments for 305 incomplete profiles with the Figure Eight crowdsourcing platform. The workers were instructed to choose `None of the above' when no choice seemed appropriate, and `I can not decide' when all values seemed equally intuitive. We picked reliable, US-based workers, and ensured US minimum wage (\$7.25) payment.

Given that there is no `correct' answer for a profile and our annotators' guesses are influenced by their subjective experiences, it is reasonable that they have a different intuition in some cases. Hence, the relatively low mean \citeauthor{krippendorff1980content} (\citeyear{krippendorff1980content}) alpha agreement per property (0.203) is not entirely surprising. We note that the agreement on the high-entropy attributes is typically lower, but tends to increase as more facets were provided. Overall, the annotators chose a concrete value rather than being indecisive (`I can not decide') for the low-entropy more often than the high-entropy facets. When more properties were provided, the frequency of indecisiveness on the high-entropy facets declined.

%We performed no further postprocessing of the data.

\begin{table}[!t]
\centering
\small
\caption{Human evaluation results per attribute: number of values ($v_i$), entropy ($H_i$), normalized entropy ($H_i\prime$), mean judgments entropy ($J_i$), divergences of: MFV, NB, and AE. }
\begin{tabular}{r | c c c c | c c c }
\textbf{attribute} & \textbf{$v_i$} & \textbf{$H_i$} & \textbf{$H_i\prime$} & \textbf{$J_i$} & \textbf{MFV} & \textbf{NB} & \textbf{AE} \\ \toprule 
cent. of birth & 5 & 0.40 & 0.92 & $10\textsuperscript{-8}$ & 0.13 & \textbf{0.12} & \textbf{0.12} \\ \midrule
religion & 4 & 0.63 & 1.26 & $10\textsuperscript{-10}$ & \textbf{0.05} & 0.09 & 0.06 \\ \midrule
sex or gender & 2 & 0.70 & 0.70 & $10\textsuperscript{-14}$ & 0.04 & \textbf{0.02} & \textbf{0.02} \\ \midrule
place of death & 8 & 0.80 & 2.40 & 0.05 & 0.51 & 0.20 & \textbf{0.16} \\ \midrule
lifespan range & 10 & 0.81 & 2.68 & 0.02 & 0.29 & \textbf{0.09} & \textbf{0.09} \\ \midrule
place of birth & 8 & 0.83 & 2.48 & 0.01 & 0.39 & 0.26 & \textbf{0.24} \\ \midrule
work location & 10 & 0.84 & 2.80 & 0.03 & 0.49 & \textbf{0.28} & 0.30 \\ \midrule
occupation & 9 & 0.92 & 2.90 & 0.06 & 0.37 & 0.36 & \textbf{0.32}\\ \midrule
educated at & 9 & 0.92 & 2.91 & 0.06 & 0.39 & 0.25 & \textbf{0.23} \\ \midrule
political party & 2 & 1.00 & 1.00 & 0.02 & 0.17 & \textbf{0.06} & \textbf{0.06} \\  \bottomrule
\end{tabular}
\label{tab:human_results}
\end{table}

\subsection{Results}

When evaluating, `None of the above' was equalized to any value outside of the most frequent 10, and `I can not decide' to a $(1/N)$-th vote for each of the values. 
The human judgments per profile were combined in a single distribution, and then compared to the system distribution by using Jansen-Shannon divergence (\textit{JS-divergence}).%to measure the difference between our system judgment and the data obtained by the crowd.
\footnote{We considered the following metrics: JS-divergence, JS-distance, KL-divergence, KL-divergence-avg, KL-divergence-max, and cosine distance~\cite{mohammad2012distributional}. The agreement was very high, the lowest Spearman correlation being 0.894.} We evaluate the profiles generated by our AE and the baselines; EMB could not be tested on this data since most inputs do not have a corresponding Wikipedia page and pre-trained embeddings.

%\todo{piek: so what does this mean? discuss results in more detail}
The divergence between our AE system and the human judgments was mostly lower than that of the baselines (Table \ref{tab:human_results}). The divergences for any system have a strong correlation with (normalized) entropy, %, and to a lesser extent with normalized entropy, 
confirming our previous observation that high-entropy attributes pose a greater challenge. %Typically, as more features were provided, agreement rates went up; see Fig for one example. 
We also computed precision, recall, and F1-score between the classes suggested by our system and by the annotators, and observed that it correlates inversely with the entropy in the data ($H_i$), as well as the entropy of the human judgments ($J_i$).
We refer the reader to our next work (\textit{anonymized, in preparation}) for further details on the obtained results.

The results show that our AE can capture human-like expectations better than the two baselines, and that mimicking human profiling is more difficult when the entropy is higher. While parameter tuning and algorithmic inventions might improve the profiling accuracy further, it is improbable that profiles learned on factual data would ever equal human performance. Some human expectations are rather culturally projected, and do not correspond to episodic facts. Future work should seek novel solutions for this challenge.

\section{Limitations of Profiling by NNs}
\label{sec:limitations}

%Whenever rich schema is not available, traditional Knowledge representation methods for inference can not be applied. However, 
Our experiments show the natural power of neural networks to generalize over knowledge and generate profiles from data independent of schema availability. Techniques like dropout and oversampling further boost their ability to deal with missing or underrepresented values. Ideally these profiling machines can be included in an online active representation system to create profiles on the fly, while their modularity allows easy retraining in the background when needed.

Still, it is essential to look critically beyond the accuracy numbers, and identify the strengths and weaknesses of the proposed profiling methods.  Limitations include: 
\begin{enumerate*}
\item \textbf{continuous values}, such as numbers (e.g., age) or dates (e.g., birth date), need to be categorized before being used in an AE;\footnote{We obtained \textit{lifespan} and \textit{century of birth} from birth and death dates, see our github page $<anonymizedlink>$ for details.}
%\item AEs can not natively deal with \textbf{continuous values}, such as numbers (e.g., age) or dates (e.g., birth date). To include this data in an AE, one has to categorize it first;\footnote{We obtained \textit{lifespan} and \textit{century of birth} from birth and death dates, see our github page $<anonymizedlink>$ for details.}
%\item AEs are trained to classify categorical values and cannot natively deal with \textbf{continuous values}, such as numbers (e.g., age) or dates (e.g., birth date), for which expectations as intervals of normal values are very valuable. To include continuous data in an AE, one has to categorize it in a preprocessing step;\footnote{We obtained \textit{lifespan} and \textit{century of birth} from birth and death dates, see our github page $<anonymizedlink>$ for details.}%\footnote{The last two facets were categorized in a preprocessing step, }

%We have done this for the attributes \textit{lifespan range} and \textit{century of birth}, generated from the attributes \textit{date of birth} and \textit{death date}.}
\item AE cannot natively handle \textbf{multiple values} (e.g., people with dual nationality). We currently pick a single value from a set based on frequency; %  Our workaround was a simple heuristic of choosing the value with lowest ID from the set of multiple values.
%\qizhe{The ID is to choose frequent/common entities}
\item as noted, we applied dropout and oversampling mechanisms to reinforce \textbf{sparse attributes}, but these remain problematic;
%\item as noted, our training requires tricks to reinforce \textbf{sparse attributes}, especially when there is no value for an attribute in a batch of data. Dropout and oversampling mechanisms address this issue, but only partially;
%\qizhe{The problem is the data-sparsity}
%however, the methods sometimes still predict an empty value as a second or third most probable option. 
\item it remains unclear \textbf{which aspects of the knowledge} are captured by our neural methods, especially by the EMB model whose embeddings abstract over the bits of knowledge. More insight is required to explain some differences we observed on individual facets. %This especially concerns the EMB model, as its embeddings abstract over the individual facets. %We are thus currently unable to identify the source of difference in accuracy between the two neural architectures on stereotyping individual facets like \textit{religion} or \textit{work location}. 
\end{enumerate*}

%Our qualitative analysis shows that our machine can mostly understand spatial dependencies (e.g. that politicians in US tend to come from Yale or Harvard, while the ones in the Netherlands from University of Amsterdam or Radboud University Nijmegen). This seems to be captured to a lesser extent for the temporal aspect of knowledge.

%\input{hunger_nlp.tex}

\section{Conclusion and Future Work}
\label{sec:conclusion}

%Profiling is a cognitive skill implicitly employed by people as a mechanism to fill gaps in knowledge with expectations. 
Inspired by the functions of profiles in human cognition, in this paper we defined the task of profiling over incomplete knowledge in the belief that AI systems that include an accessible profiling component can naturally fill knowledge gaps with assumed expectations.
KBC and other existing tasks can not be used for this purpose, since they focus on predicting concrete missing facts (exact age or location) rather than distributions over ranges of values.
%While the phenomenon has been well-researched in fields like social psychology, it is surprisingly unaddressed in AI: no computational system has the ability to create profiles and expectations based on partial knowledge. In this paper, we formulate this task, 
We described two profiling machines based on state-of-the-art neural network techniques. We demonstrated their skills in comparison to human judgments as well as existing instantial data.

Data scarcity is (unfortunately) a rather prevalent phenomenon, with most instances being part of the Zipfian long tail. Applications in NLP and IR suffer from \textbf{hunger for knowledge}, i.e. a lack of information on the tail instances in knowledge bases and in communication.
%Hunger for knowledge as the next step in the cognitive journey of AI, which would rely on , to fill gaps 
%When very little is known, a lot can be inferred. According to the pragmatic principles of human communication, people rely on their expectations and intuition to assume most missing knowledge. 
We envision a shift in the process of creation of AI knowledge bases to incorporate human skills such as profiling, type inference, etc. Knowledge bases built on cognitive grounds would be able to natively address (at least) three standing problem areas:
\begin{enumerate*}
\item \textbf{scarcity of episodic knowledge}, prominent both in knowledge bases and in communication; % - this scenario occurs when the episodic data in text and directly in the background knowledge source is too scarce to come up with a reliable interpretation. %For instance, stereotyping that basketball players come often from North Carolina and scientists from Carnegie Mellon University would help our machine to generate expectations whenever the newspaper is talking about professionals that studied at some of these universities.
\item \textbf{unresolved ambiguity} in communication, when the available knowledge is not necessarily scarce, yet prior expectations could lead to more reliable disambiguation; %As an example, consider a news article published in 1918, discussing the presidency candidature of Mr. Ford. Both Gerald Ford and Henry Ford have ran for presidents, and both of them were alive when this article was published. However, it is much more stereotypical that the president is Henry Ford, who was 55 years old at the moment of writing of that article, and not the 5-year-old Gerald Ford.
\item \textbf{anomaly detection}, when a seemingly reliable machine interpretation is counter-intuitive and anomalous with respect to our expectations.%. For instance, it is very stereotypical that the members of a Christian party belong to one of the Christian religions. If the system interprets an atheist or a Buddhist as a member of this party, that would collide with our stereotypes, triggering an anomaly.
\end{enumerate*}

\bibliography{main}
\bibliographystyle{aaai}

\end{document}